\documentclass[runningheads,a4paper]{llncs}

\usepackage{amsmath, amsfonts, amssymb, algorithmic}
\usepackage{graphicx}

\usepackage{url}
\urldef{\mailsa}\path|everyan@cs.unm.edu, yaojia.zhu@gmail.com|
\urldef{\mailsb}\path|jrouquie@gmail.com, moore@santafe.edu|

\newcommand{\keywords}[1]{\par\addvspace\baselineskip
\noindent\keywordname\enspace\ignorespaces#1}
\newcommand{\like}{\mathcal{L}}
\newcommand{\MI}{{\rm MI}}
\renewcommand{\AA}{{\rm AA}}
\newcommand{\abs}[1]{\left| #1 \right| }
\DeclareMathOperator*{\argmax}{{\rm argmax}}

\begin{document}

\mainmatter  

\title{Active Learning for\\ Hidden Attributes in Networks}

\titlerunning{Active Learning for Network Attributes}

%
%
\author{Xiaoran Yan \inst{1}%
\and Yaojia Zhu \inst{1} \and Jean-Baptiste Rouquier \inst{2}\and Cristopher Moore \inst{1}\inst{3}}
\authorrunning{Xiaoran Yan, Yaojia Zhu et al.}

\institute{University of New Mexico, Albuquerque, NM, USA
\and
Complex Systems Institute Rh\^{o}ne-Alpes, ENS Lyon, France
\and
Santa Fe Institute, Santa Fe, NM, USA
\mailsa\\
\mailsb\\
}

%
%

\toctitle{Lecture Notes in Computer Science}
\tocauthor{Authors' Instructions}
\maketitle

\begin{abstract}
In many networks, vertices have hidden attributes, or types, that are correlated with the network’s topology. If the topology is known but
these attributes are not, and if learning the attributes is costly, we need a method for choosing which vertex to query in order to learn as much as possible about the attributes of the other vertices.  We assume the network is generated by a stochastic block model, but we
make no assumptions about its assortativity or disassortativity. We choose which vertex to query using two methods: 1) maximizing the
mutual information between its attributes and those of the others (a well-known approach in active learning) and 2) maximizing the average
agreement between two independent samples of the conditional Gibbs distribution. Experimental results show that both these methods do
much better than simple heuristics. They also consistently identify certain vertices as important by querying them early on.

\end{abstract}

\section{Introduction}

Suppose we have a network, represented by a graph $G=(V,E)$ with $n$ vertices.  Suppose further that each vertex $v$ has a type $t(v) \in \{1,\ldots,k\}$, representing the value of some hidden attribute that takes $k$ different values.  We are given the graph $G$, and our goal is to learn the types $t(v)$.  One way we might do this is to assume that $G$ is generated by some probabilistic model, in which its topology is correlated with these types.

The simplest such model, although by no means the only one to which our methods could be applied, is a stochastic block model.  Here we assume that each pair of vertices $u,v$ have an edge between them with a probability $p_{t(u),t(v)}$, and that these events are independent.  Given an assignment $t:V \to \{1,\ldots,k\}$ of types to vertices, and a $k \times k$ matrix of probabilities $p_{ij}$, the likelihood of generating $G$ in this model is
\begin{align}
\like(G \,|\, t,p)
&= \left( \prod_{(u,v) \in E} p_{t(u),t(v)} \right) \left( \prod_{(u,v) \notin E} (1-p_{t(u),t(v)}) \right)
\nonumber \\
&= \prod_{i,j=1}^k p_{ij}^{e_{ij}} (1-p_{ij})^{n_i n_j - e_{ij}} \; ,
\label{eq:like-tp}
\end{align}
where $n_i = \abs{ \{ v \in V : t(v) = i \} }$ is the number of vertices of type $i$, and $e_{ij} = \abs{ \{ (u,v) \in E : t(u) = i, t(v) = j \} }$ is the number of edges from vertices of type $i$ to vertices of type $j$.  Note that~\eqref{eq:like-tp} assumes that edges are directed, and allows self-loops.  We can disallow self-loops, or make the edges undirected, by replacing $n_i^2$ with $n_i (n_i-1)$ or ${n_i \choose 2}$ respectively, and/or taking the product over pairs of types $i,j$ with $i \le j$.

We do not assume that $p_{ij}$ takes one value when $i=j$ and a smaller value when $i \ne j$.  In other words, we do not assume an assortative community structure, where vertices are more likely to be connected to other vertices of the same type.  Nor do we require that $p_{ij} = p_{ji}$, since the directed nature of the edges may be important.  For example, herbivores eat plants, but the reverse is usually not the case.  This kind of stochastic block model is well-known in the sociology and machine learning communities (e.g.~\cite{wang_stochastic_1987,rosvall_information-theoretic_2007,airoldi_mixed_2008,hofman_bayesian_2008}) and has also been used in ecology to identify groups of species in food webs~\cite{letters_food_2009}.  

Since we are interested in finding the labels $t$ of the nodes, we integrate over the parameters $p_{ij}$ of the block model, 
in order to obtain the likelihood of $G$ given $t$.  If we assume a prior, in which the $p_{ij}$ are independent this integral factorizes over the product~\eqref{eq:like-tp}.  In particular, if each $p_{ij}$ is chosen uniformly from $[0,1]$, we have
\begin{align}
\like(G \,|\, t)
&= \iiint_0^1 {\rm d}\{p_{ij}\} \,\like(G \,|\, t,p) \nonumber \\
&= \prod_{i,j=1}^k \int_0^1 {\rm d}p_{ij} \,p_{ij}^{e_{ij}} (1-p_{ij})^{n_i n_j - e_{ij}} \nonumber \\
&= \prod_{i,j=1}^k \frac{1}{(n_i n_j + 1) {n_i n_j \choose e_{ij}}} \; . \label{eq:gibbs}
\end{align}
Of course, we could easily assume some other prior on $[0,1]$ for the $p_{ij}$, such as a beta distribution, and then optimize its parameters, but here we will stick to~\eqref{eq:gibbs} for its simplicity.  If we assume a uniform prior over the assignments $t$, then Bayes' rule gives them a Gibbs distribution
\begin{equation}
\label{eq:gibbs2}
P(t) = P(t \,|\, G) \propto \like(G \,|\, t) \; .
\end{equation}
Note that~\eqref{eq:gibbs} is maximized when, for each pair of types, $e_{ij}$ is close to $0$ or to $n_i n_j$.  In other words, the most likely assignments are those where, for each pair of types $i,j$, pairs of vertices of types $i$ and $j$ are either mostly connected or mostly unconnected.

An alternate approach is to assume that the $p_{ij}$ take their maximum likelihood values
\begin{equation}
\hat{p}_{ij} = \argmax_p \,\like(G \,|\, t,p) = e_{ij} / n_i n_j \; ,
\end{equation}
and set $\like(G \,|\, t) = \like(G \,|\, t,\hat{p})$.  This approach was used, for instance, for a hierarchical block model in~\cite{clauset_hierarchical_2008}.  When $k$ is fixed and the $n_i$ are large, this will give results similar to~\eqref{eq:gibbs}, since the integral over $p$ is tightly peaked around $\hat{p}$.  However, for any particular finite graph it makes more sense, at least to a Bayesian, to integrate over the $p_{ij}$, since they obey a posterior distribution rather than taking a fixed value.  Moreover, averaging over the parameters as in~\eqref{eq:gibbs} discourages over-fitting, since the average likelihood goes down when we increase $k$ and hence the volume of the parameter space.  This should allow us to determine $k$ automatically, although in this paper we set $k$ by hand.

We emphasize, however, that the approaches to active learning we discuss below are not tied to this particular type of block model.  They can be adapted to a wide range of other probabilistic models in which topology is correlated with hidden attributes of the vertices.

We note that Bilgic and Getoor have discussed ways to use network relationships to improve active learning about vertices~\cite{bilgic_link-based_2009}, and that Hanneke and Xing~\cite{hanneke_network_2009} have studied active learning for learning network topology.  In contrast to~\cite{hanneke_network_2009}, we assume that the network topology is known, but that the types of the vertices are not.

\section{Active Learning}

In the active learning setting, the algorithm can learn the type of any given vertex, but at a cost---say, by devoting resources in the laboratory or the field.  Since these resources are limited, it has to decide which vertex to query.  Its goal is to query a small set of vertices, and use their types to make good guesses about the types of the remaining vertices.  

One natural approach (see, e.g., MacKay~\cite{mackay_information-based_1992} or Guo and Greiner~\cite{guo_optimistic_2007}) is to query the vertex $v$ with the largest mutual information (MI) between its type $t(v)$ and the types $t(G \setminus v)$ of the other vertices.  We can write this as the difference between the entropy of $t(G \setminus v)$ and its conditional entropy given $t(v)$,
\begin{equation}
\MI(v) = I(v ; G \setminus v) = H(G \setminus v) - H(G \setminus v \,|\, v) \; . 
\end{equation}
Here $H(G \setminus v \,|\, v)$ is the entropy, averaged over $t(v)$ according to the marginal of $t(v)$ in the Gibbs distribution, of the joint distribution of $t(G \setminus v)$ conditioned on $t(v)$.  In other words, $\MI(v)$ is the expected decrease in the entropy of $t(G \setminus v)$ that will result from learning $t(v)$.  Since the mutual information is symmetric, we also have
\begin{equation}
\MI(v) = I(v ; G \setminus v) = H(v) - H(v \,|\, G \setminus v) \; , 
\end{equation}
where $H(v)$ is the entropy of the marginal distribution of $t(v)$, and $H( v \,|\, G \setminus v)$ is the entropy, on average, of the distribution of $t(v)$ conditioned on the types of the other vertices.  Thus a good vertex to query is one about which we are quite uncertain, so that $H(v)$ is large---but which is strongly correlated with other vertices, so that $H(v \,|\, G \setminus v)$ is small.

We estimate these entropies by sampling from the space of assignments according to the Gibbs distribution.
Specifically, we use a single-site heat-bath Markov chain.  At each step, it chooses a vertex $v$ uniformly from among the unqueried vertices, and chooses $t(v)$ according to the conditional distribution proportional to $\like(G \,|\, t)$, assuming that the types of all other vertices stay fixed.  In addition to exploring the space, this allows us to collect a sample of the conditional distribution of the chosen vertex $v$ and its entropy.  Since $H( v \,|\, G \setminus v)$ is the average of the conditional entropy, and since $H(v)$ is the entropy of the average conditional distribution, we can write
\begin{equation}
I(v ; G \setminus v) = - \sum_{i=1}^k \overline{P_i} \ln \overline{P_i} + \overline{ \sum_{i=1}^k P_i \ln P_i} \; , 
\end{equation}
where $P_i$ is the probability that $t(v)=i$.

We offer no theoretical guarantees about the mixing time of the heat-bath Markov chain, and it is easy to see that there are families of graphs and values of $k$ for which it grows exponentially with $n$.  For instance, if $G$ is an Erd\H{o}s-R\'enyi random graph $G(n,1/2)$, in which each pair of vertices is independently connected with probability $1/2$, and if $k=2$, it takes $2^{\Omega(n^2)}$ steps on average to switch from a state where most vertices are of type $1$ to one where most are of type $2$, since the ``bottleneck'' states where half the vertices are of each type have total probability $2^{-\Omega(n^2)}$.  However, for the real-world networks we have tried so far, the Markov chain appears to converge to equilibrium, and give good estimates of $\MI(v)$, in a reasonable amount of time.  We also improve our estimates by averaging over many runs, each one starting from an independently random initial state.

To complete the description of the MI active learning algorithm, we say that it is in \emph{stage $j$} if it has already queried $j$ vertices.  In that stage, it estimates $\MI(v)$ for each unqueried vertex $v$, using the Markov chain to sample from the Gibbs distribution conditioned on the types of the vertices queried so far.  It then queries the vertex $v$ with the largest MI.  We provide it with $t(v)$, and it moves on to the next stage.

Another strategy is to query the vertex that maximizes another quantity, which we call the \emph{average agreement} (AA).  Given two type assignments $t_1, t_2$, define their \emph{agreement} as the number of vertices on which they agree,
\begin{equation}
\abs{t_1 \cap t_2} = \abs{ \{ v : t_1(v) = t_2(v) \} } \; . 
\end{equation}
Since our goal is to label as many vertices correctly as possible, what we would really like to maximize is the agreement between an assignment $t_1$, drawn from the Gibbs distribution, and the correct assignment $t_2$.  But since we don't know $t_2$, the best we can do is assume that it is drawn from the Gibbs distribution as well.  If we think of $(t_1,t_2)$ as having a joint distribution, then querying $v$ would project onto the part of this distribution where $t_1(v)=t_2(v)$.  So, we define $\AA(v)$ as the expected agreement between two assignments $t_1, t_2$ drawn independently from the Gibbs distribution, conditioned on the event that they agree at $v$.  This gives us the following quantity:
\begin{equation}
\label{eq:aa}
\AA(v) = \frac{\sum_{t_1,t_2: t_1(v)=t_2(v)} P(t_1) P(t_2) \abs{ t_1 \cap t_2 }}{\sum_{t_1,t_2: t_1(v)=t_2(v)} P(t_1) P(t_2)} \, .
\end{equation}
For instance, imagine that $n=6$ and $k=2$, that vertices 1, 2, and 3 always have the same type, and that this type and the types of vertices 4, 5, and 6 are chosen uniformly and independently from $\{1,2\}$.  This gives $16$ possible assignments, each of which appears with probability $1/16$.  If $t_1$ and $t_2$ agree at 1, then they also agree at 2 and 3, and 4, 5, and 6 are each in $t_1 \cap t_2$ with probability $1/2$.  So, $\AA(1) = 3+3/2 = 9/2$.  On the other hand, if $t_1 \cap t_2$ agree at $6$, then each of the other $5$ vertices is in $t_1 \cap t_2$ with probability $1/2$, so $\AA(6) = 1+5/2 = 7/2$.  Thus we should query one of the first three vertices, because doing so will tell us the types of two other vertices as well.

We estimate $\AA(v)$ using the same heat-bath Gibbs sampler as for $\MI(v)$, except that we draw pairs of assignments $(t_1,t_2)$ independently, by starting the Markov chain at two independently chosen initial states.  We then estimate the numerator and denominator of~\eqref{eq:aa} by averaging over these pairs, giving the estimate
\begin{equation}
\label{eq:aa-est}
\AA(v)_{\rm est} = \frac{\sum_{(t_1,t_2)} \delta(t_1(v),t_2(v)) \abs{ t_1 \cap t_2 }}{\sum_{(t_1,t_2)} \delta(t_1(v),t_2(v))} \; ,
\end{equation}
where $\delta(i,j) = 1$ if $i=j$ and $0$ otherwise.  We keep track of these averages for each vertex $v$ as follows: each time we draw a pair $(t_1,t_2)$, for each $v \in t_1 \cap t_2$, we increment the numerator and the denominator of~\eqref{eq:aa-est} by $\abs{t_1 \cap t_2}$ and $1$ respectively, and for $v \notin t_1 \cap t_2$ we leave the numerator and denominator unchanged.  This gives an alternate algorithm for active learning, where in each stage we query the vertex with the largest estimated AA.

We judge the performance of these algorithms by asking, at each stage and for each vertex, with what probability the Gibbs distribution assigns it the correct type.  We can then plot, as a function of the stage $j$, what fraction of the unqueried vertices are assigned the correct type with probability at least $q$, for various thresholds $q$.

\section{Results}

\begin{figure}
\begin{center}
\includegraphics[width=0.7\textwidth]{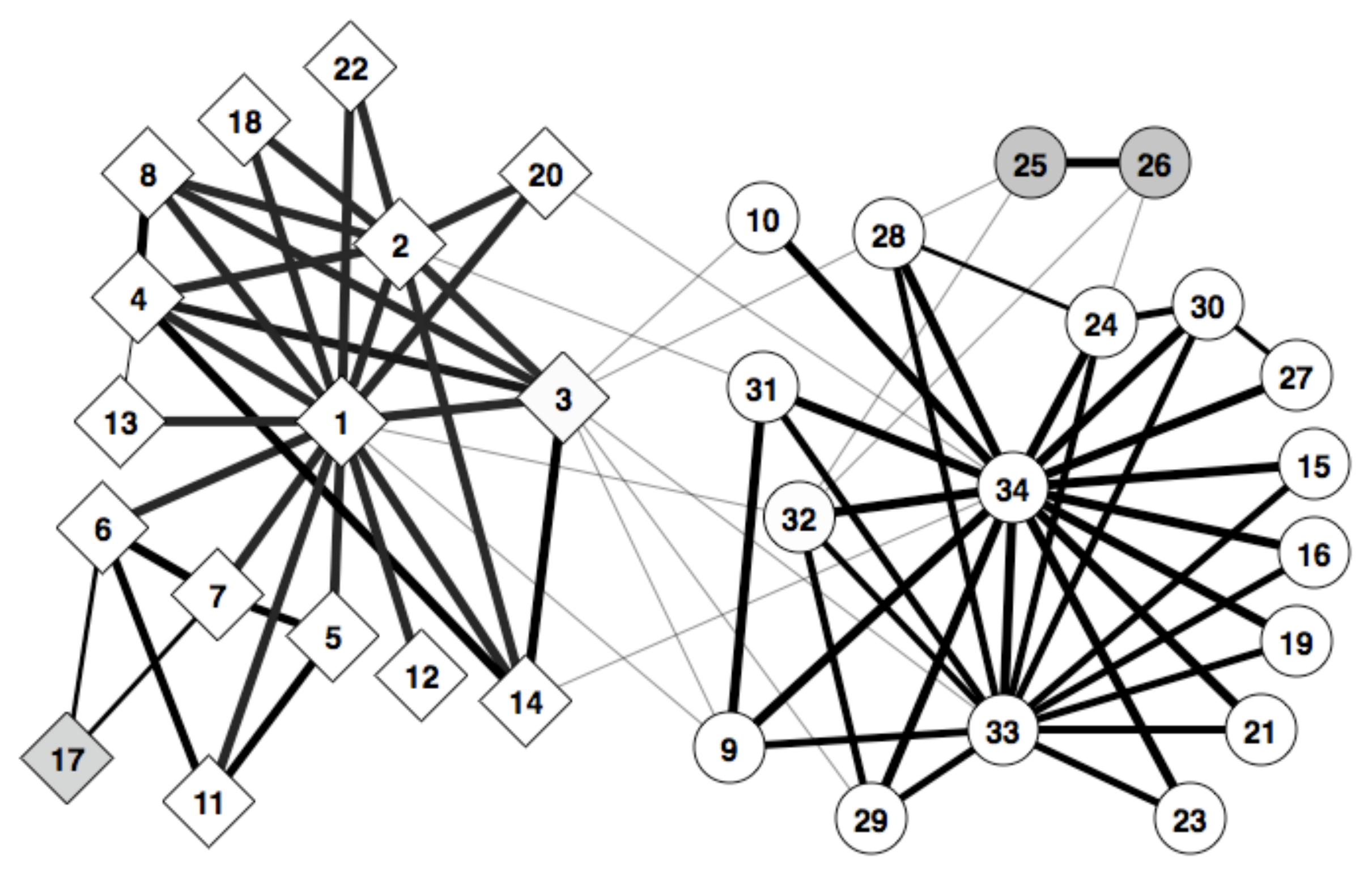}
\end{center}
\caption{Zachary's Karate Club.  Vertices $1$ and $34$ are the instructor and president, and their communities are indicated by diamonds and circles.  Shaded vertices are more peripheral, and have weaker ties to their communities.}
\label{fig:karate}
\end{figure}

\begin{figure}
\begin{center}
\includegraphics[width=0.49\textwidth]{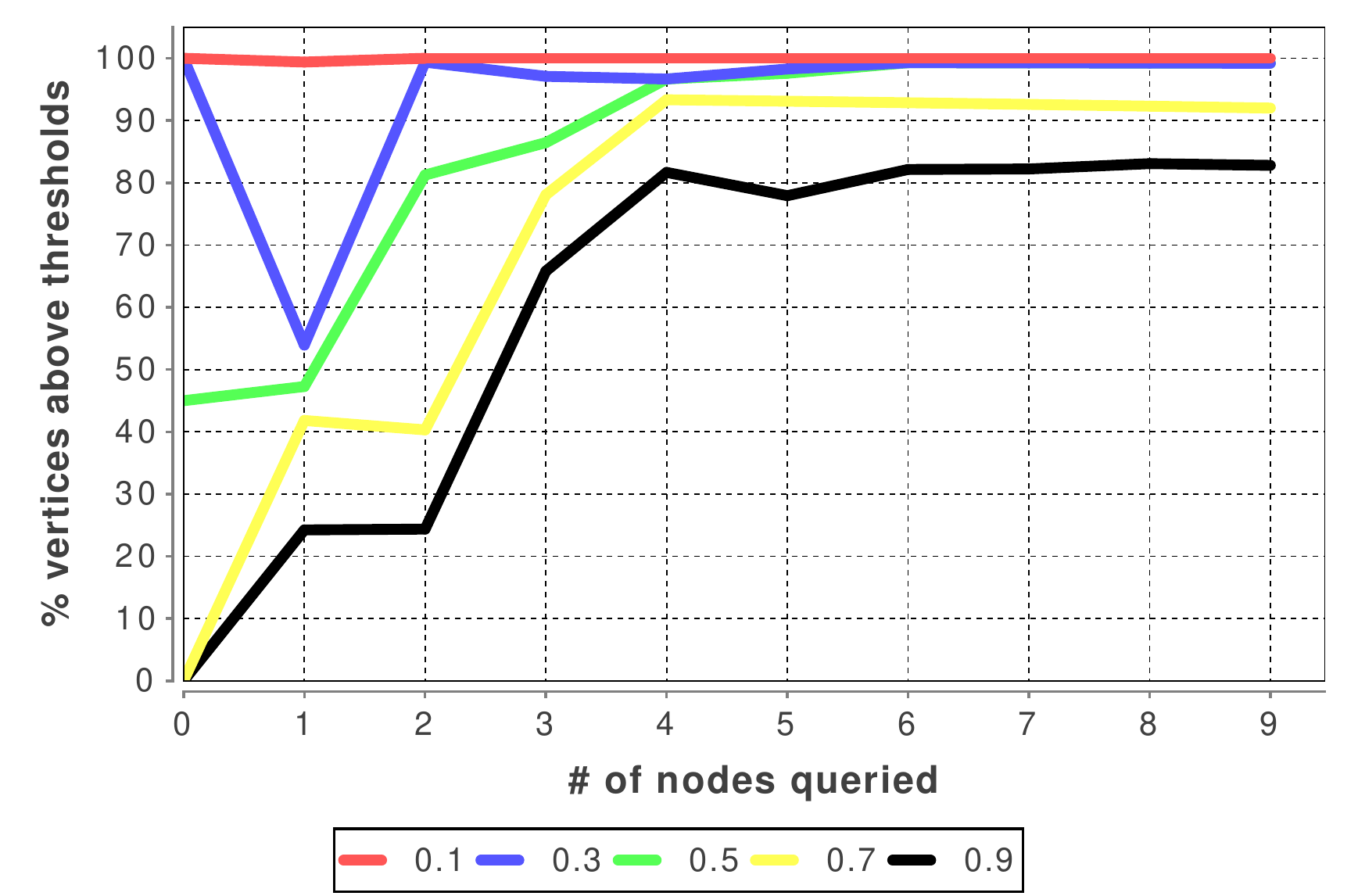}
\includegraphics[width=0.49\textwidth]{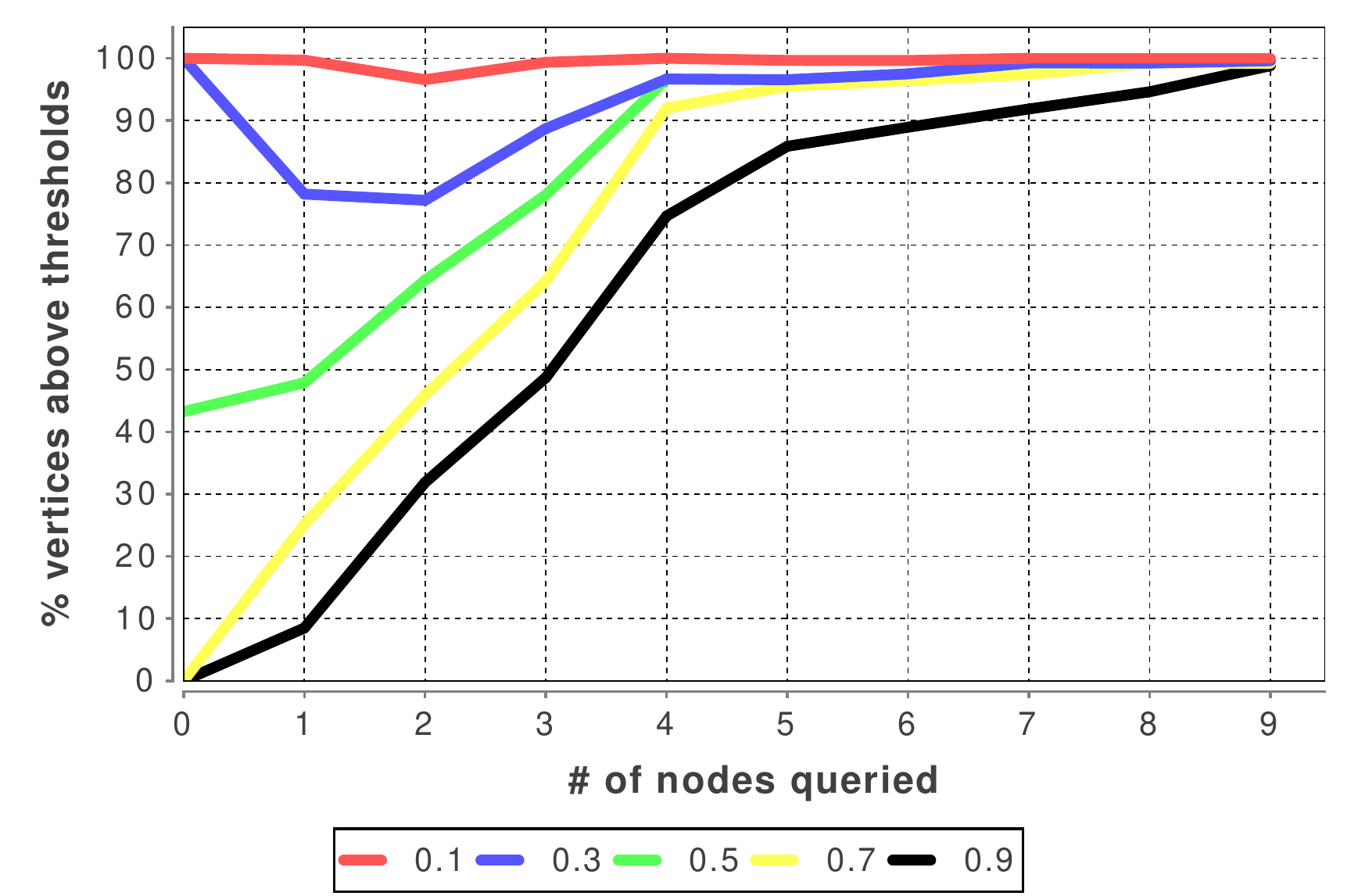}
\end{center}
\caption{Results of the active learning algorithms on Zachary's Karate Club network.  In each stage we sample the Gibbs distribution using $100$ independently chosen initial conditions, doing $2 \times 10^4$ steps of the heat-bath Markov chain for each one, and computing averages using the last $10^4$ steps.
The $y$ axis shows the fraction of vertices, other than those queried so far, which are labeled correctly by the conditional Gibbs distribution with probability at least $q$, for 
$q=0.1, 0.3, 0.5, 0.7, 0.9$. The $x$ axis is cut off after 9 queries, Fig.~\ref{fig:compare} left has the complete $0.9$ curves. 
Left, we query the vertex with the largest mutual information (MI) between it and the rest of the network.  Right, we query the vertex with the largest average agreement (AA) as defined in the text.  After querying $4$ or $5$ vertices, both methods assign the correct label to about $80\%$ of the remaining vertices with probability $0.9$ or greater.  The AA algorithm performs somewhat better, with the accuracy quickly converging to $100\%$ as it queries more vertices.}
\label{fig:learnkarate}
\end{figure}

We tested the MI and AA algorithms on Zachary's Karate Club~\cite{zachary_information_1977}, shown in Fig.~\ref{fig:karate}.  This is a social network consisting of $34$ members of a karate club, where edges represent friendships.  The club split into two factions, one centered around the instructor (vertex 1) and the other around the club president (vertex 34), each of which formed their own club.  The network is highly associative, with a high density of edges within each faction and a low density of edges between them.

It is no surprise that, after querying just four or five vertices, both algorithms succeed in correctly identifying the types of most of the remaining vertices---i.e., to which faction they belong---with high accuracy.  The AA algorithm performs slightly better, achieving an accuracy close to $100\%$ after nine queries.  These results are shown in Fig.~\ref{fig:learnkarate}.  Of course, this network is quite small, and there are many community structure algorithms that identify the two factions with perfect or near-perfect accuracy; see e.g.~\cite{porter_communities_2009,s._fortunato_community_2009} for reviews.

\begin{figure}
\begin{center}
\includegraphics[width=\textwidth]{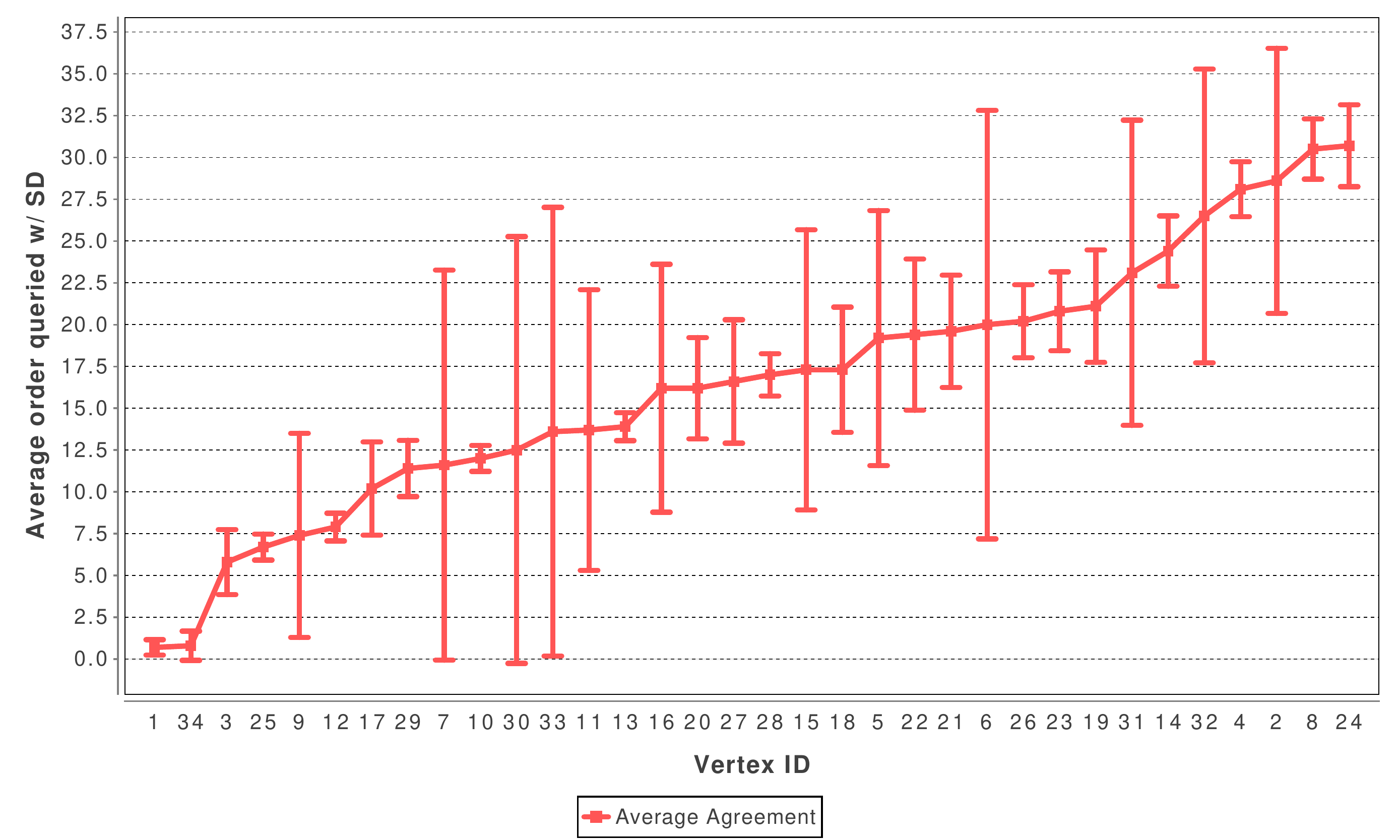}
\includegraphics[width=\textwidth]{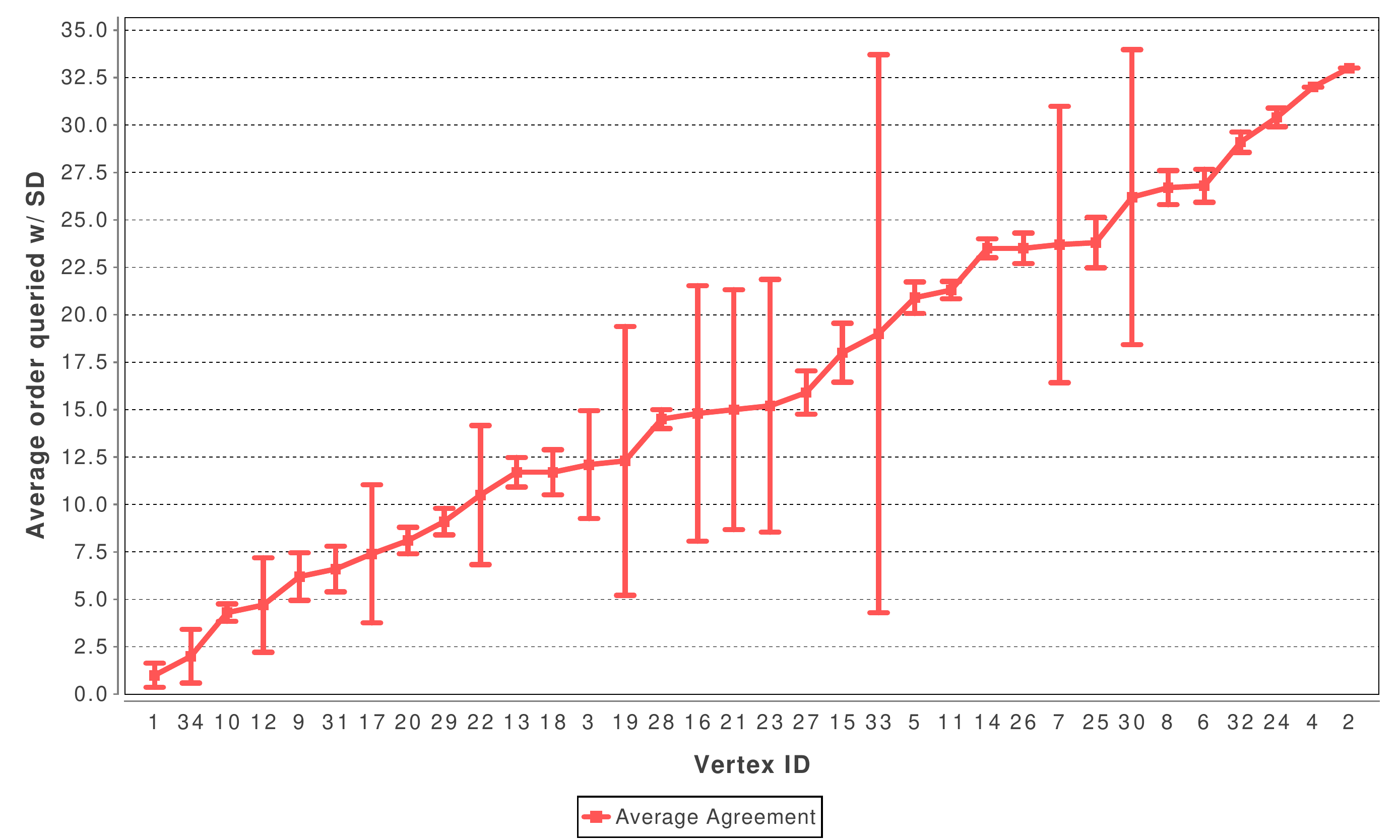}
\end{center}
\caption{The order in which the active learning algorithms query vertices in Zachary's Karate Club network, averaged over $10$ independent runs of each algorithm.  Error bars show the standard deviation.  Both algorithms start by querying vertices $1$ and $34$, which are central to their respective communities, and then query vertices at the boundary between the two communities.}
\label{fig:querykarate}
\end{figure}

Perhaps more interesting is the order in which these algorithms choose to query the vertices.  In Fig.~\ref{fig:querykarate}, we sort the vertices in order of the average stage at which they are queried.  Both algorithms start by querying the two central vertices, the instructor and the president.  They then query vertices such as 3, 9, and 10, which lie at the boundary between the two communities.  At that point, the algorithms ``understand'' that the network consists of two assortative communities, and the boundary between the communities is clear.  The last vertices to be queried are those such as 2, 4, and 24, which lie deep inside their communities, so that their types are not in doubt.  It is not clear why the AA algorithm performs better, but from this small experiment, it seems that it places a lower priority on peripheral vertices, such as 25, than the MI algorithm does.

We also examined a food web of $488$ species in the Weddell Sea, in the Antarctic~\cite{brose_body_2005,NewEntry1}.  This data set is very rich, but we focus on two particular attributes---the feeding type, and the part of the environment, or habitat, in which the species lives.  The feeding type takes $k=5$ values, namely herbivorous, carnivorous, omnivorous, detritivorous, or a primary producer.  The habitat attribute takes $k=5$ values, namely pelagic, benthic, benthopelagic, demersal, and land-based.

\begin{figure}
\begin{center}
\includegraphics[width=0.49\textwidth]{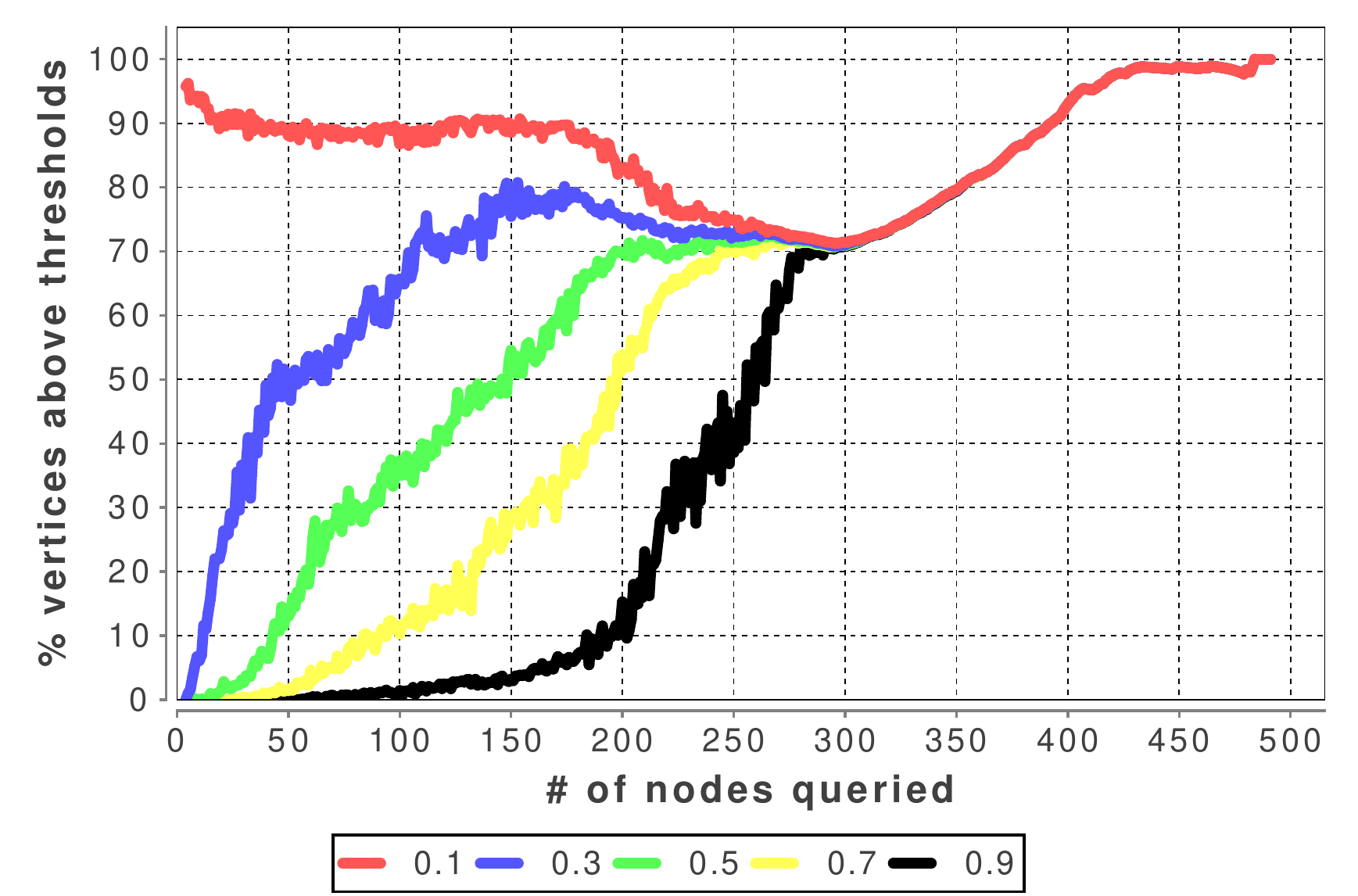}
\includegraphics[width=0.49\textwidth]{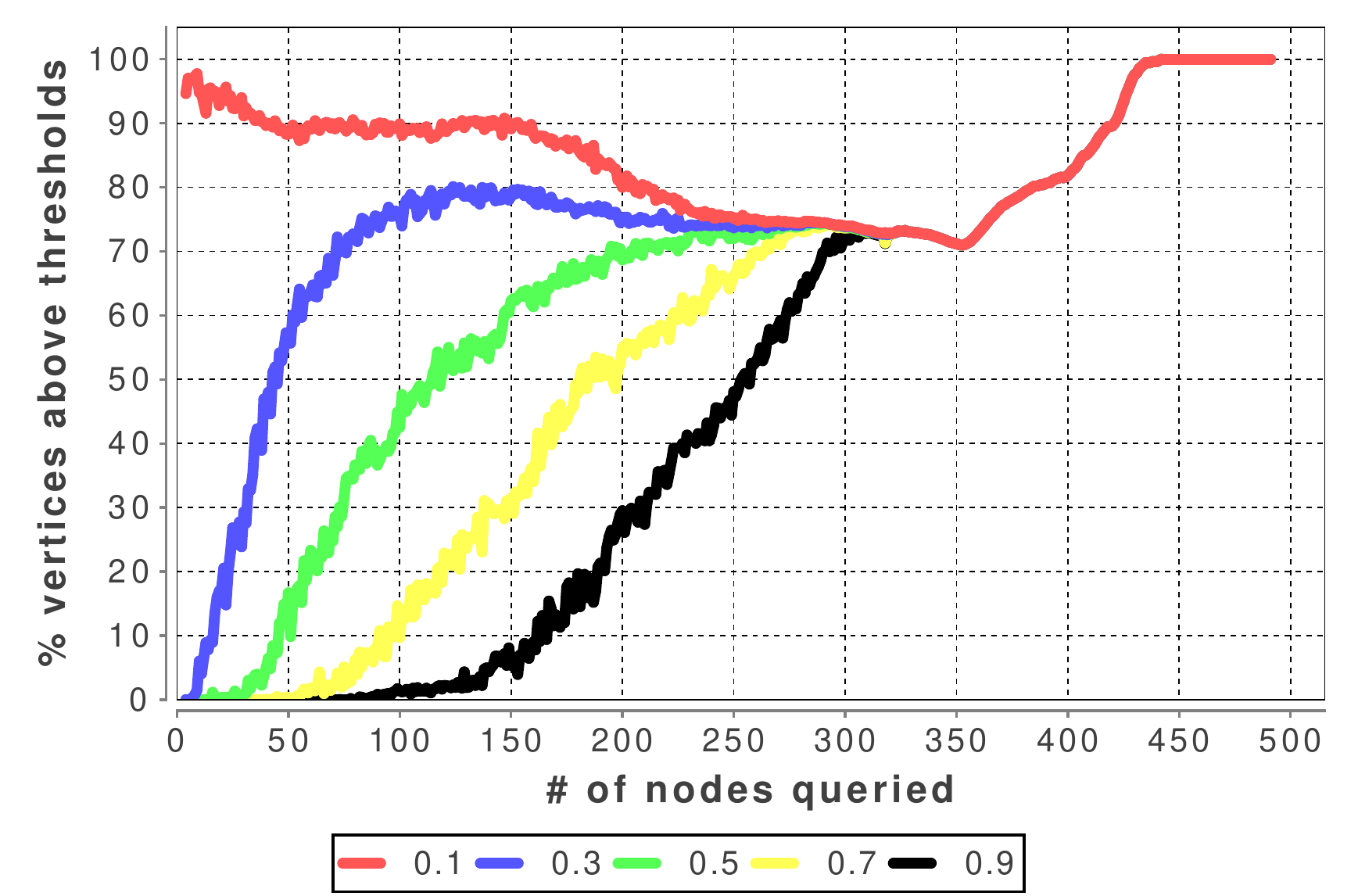}
\includegraphics[width=0.49\textwidth]{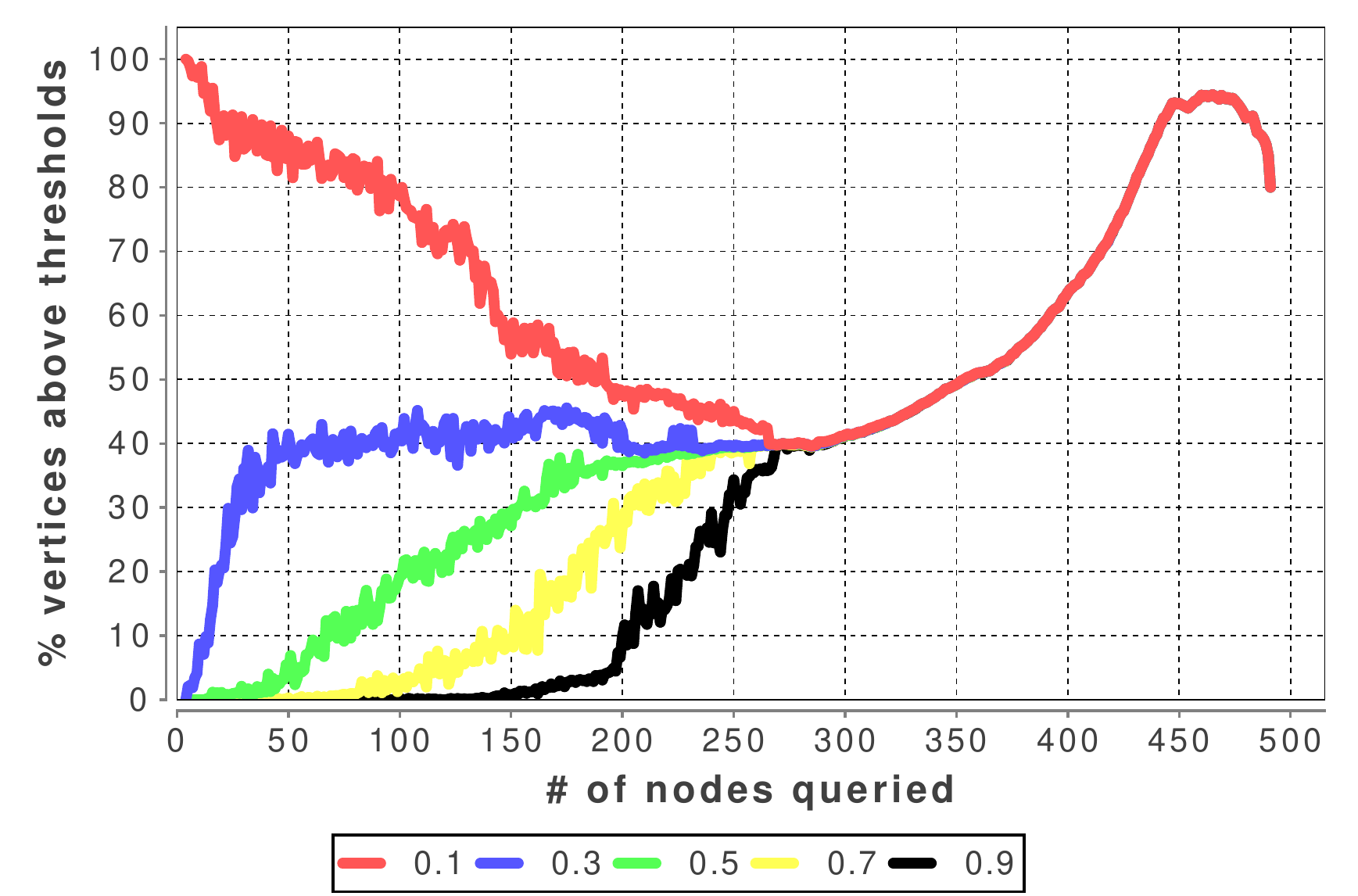}
\includegraphics[width=0.49\textwidth]{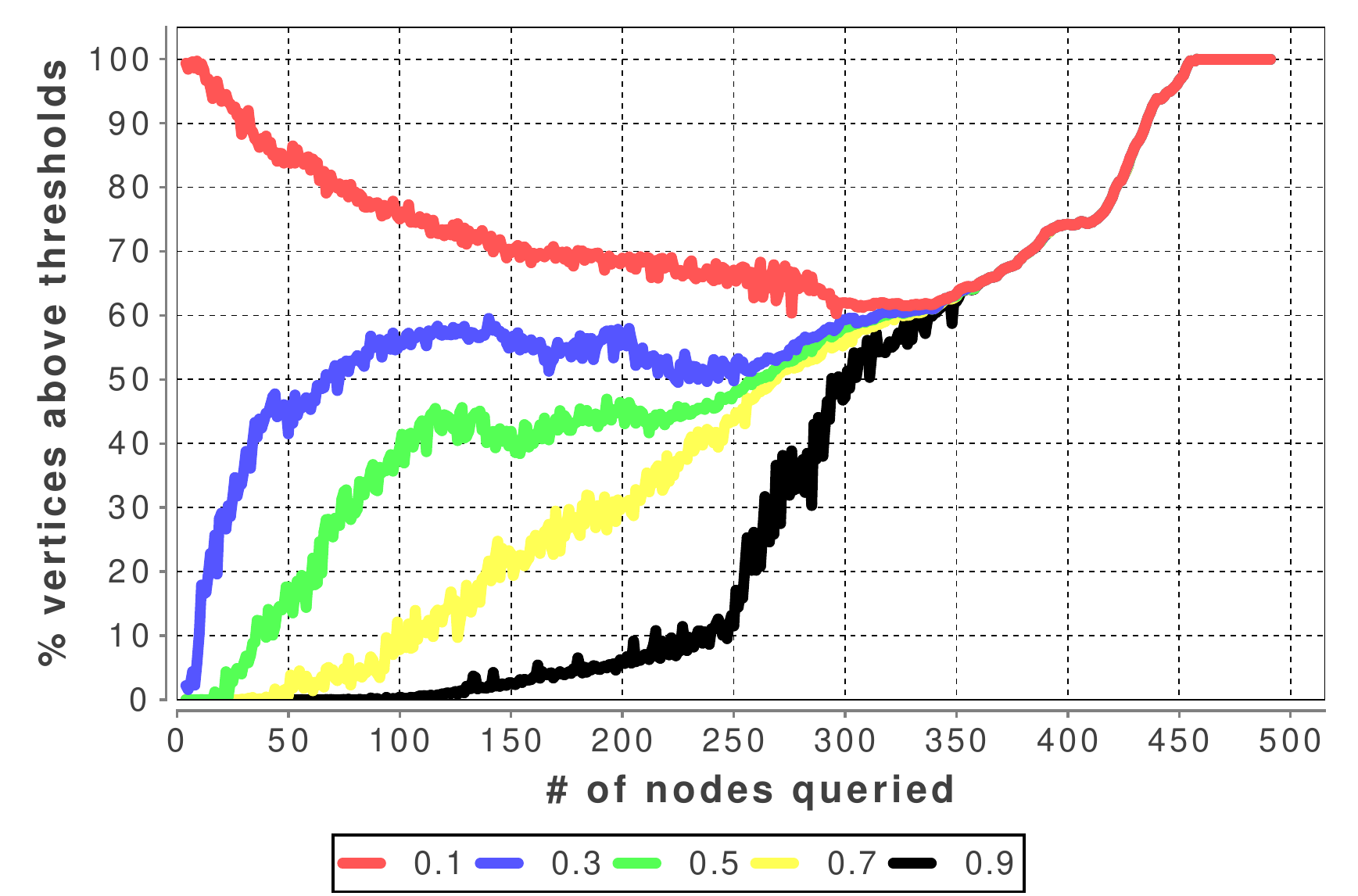}
\end{center}
\caption{Results for the Weddell Sea food web, averaged over 10 runs of each algorithm.  In each stage we sample the Gibbs distribution using $100$ independently chosen initial conditions, doing $5\times 10^4$ steps of the heat-bath Markov chain for each one, and computing averages using the last $2.5 \times 10^4$ steps.   The $y$ axis shows the fraction of vertices, other than those queried so far, which are labeled correctly by the conditional Gibbs distribution with probability at least $q$, for $q=0.1, 0.3, 0.5, 0.7, 0.9$.  The $x$ axis stops when there is only one unqueried vertex left. 
We show results for two attributes: above, the feeding type of the species, and below, the habitat in which it lives. 
After querying about half the species, both algorithms get to a stage where every species is either labeled correctly with high probability, or incorrectly with high probability.  In other words, the algorithm is confident, but wrong, about a significant fraction of the species.  Most of these are species which are poorly modeled by the stochastic block model---that is, those which would be misclassified even if one knew the types of all the other species. Left column, MI; right column, AA. }
\label{fig:learnfoodweb}
\end{figure}

We show results for both attributes in Fig.~\ref{fig:learnfoodweb}.  For feeding type, after querying half the vertices, both algorithms correctly label about $75\%$ of the remaining vertices.  For the habitat attribute, both algorithms are less accurate, although AA performs significantly better than MI.  Note that the accuracy is measured as a fraction of the un-queried vertices.  It can decrease, for instance, if we query ``easy'' vertices early on, so that ``hard'' vertices form a larger fraction of the remaining ones.

Fig.~\ref{fig:learnfoodweb} also shows that both algorithms arrive at a stage at which they are either right most of the time, or wrong most of the time, about each of the remaining vertices.  For the feeding type attribute, for instance, after the AA algorithm has queried $300$ species, it labels $75\%$ of the remaining vertices correctly with probability $90\%$---but labels the other $25\%$ correctly with probability less than $10\%$.  In other words, it has a high degree of certainty about all the vertices, but is wrong about many of them.  Its accuracy improves as it continues to query the vertices, but it doesn't achieve high accuracy on all the unqueried vertices until there are only about $60$ of them left.  For the habitat attribute, the MI algorithm gets a small fraction of the unqueried vertices wrong up until the very end of the learning process.

Thus both these algorithms find some vertices easy to classify, but others very hard.  Delving into the data, we found that, to a large extent, the blame lies not with the learning algorithms themselves, but with the stochastic block model, and its ability to model the data given this particular attribute.  For example, for the habitat attribute, these algorithms perform well on pelagic, demersal, and land-based species.  But the benthic habitat, which is the largest and most diverse, includes species with many feeding types and trophic levels.  These additional attributes have a large effect on the topology, but they are hidden from the block model in our experiments.  

As a result, more than half the benthic species are misclassified by the block model, in the sense that if we condition on the habitats of the other vertices, it believes the benthic species' most likely type is pelagic, benthopelagic, demersal, or land-based.  Specifically, 212 of the 488 species are mislabeled by the most likely block model, 94\% of them with confidence over $0.9$, even when the habitats of \emph{all} the other species are known.

To draw an analogy, if a member of the karate club was good friends with the instructor, but joined the president's club because it was close to her favorite caf\'e---and if the block model did not have access to this information---we could not expect the learning algorithm to classify her correctly until it got around to querying her.  Of course, we can also regard our algorithms' mistakes as evidence that these habitat types are not cut and dried.  Biologists are well aware that there are ``connector species'' that connect one habitat to another, and belong to some extent to both.

\begin{figure}
\begin{center}
\includegraphics[width=0.49\textwidth]{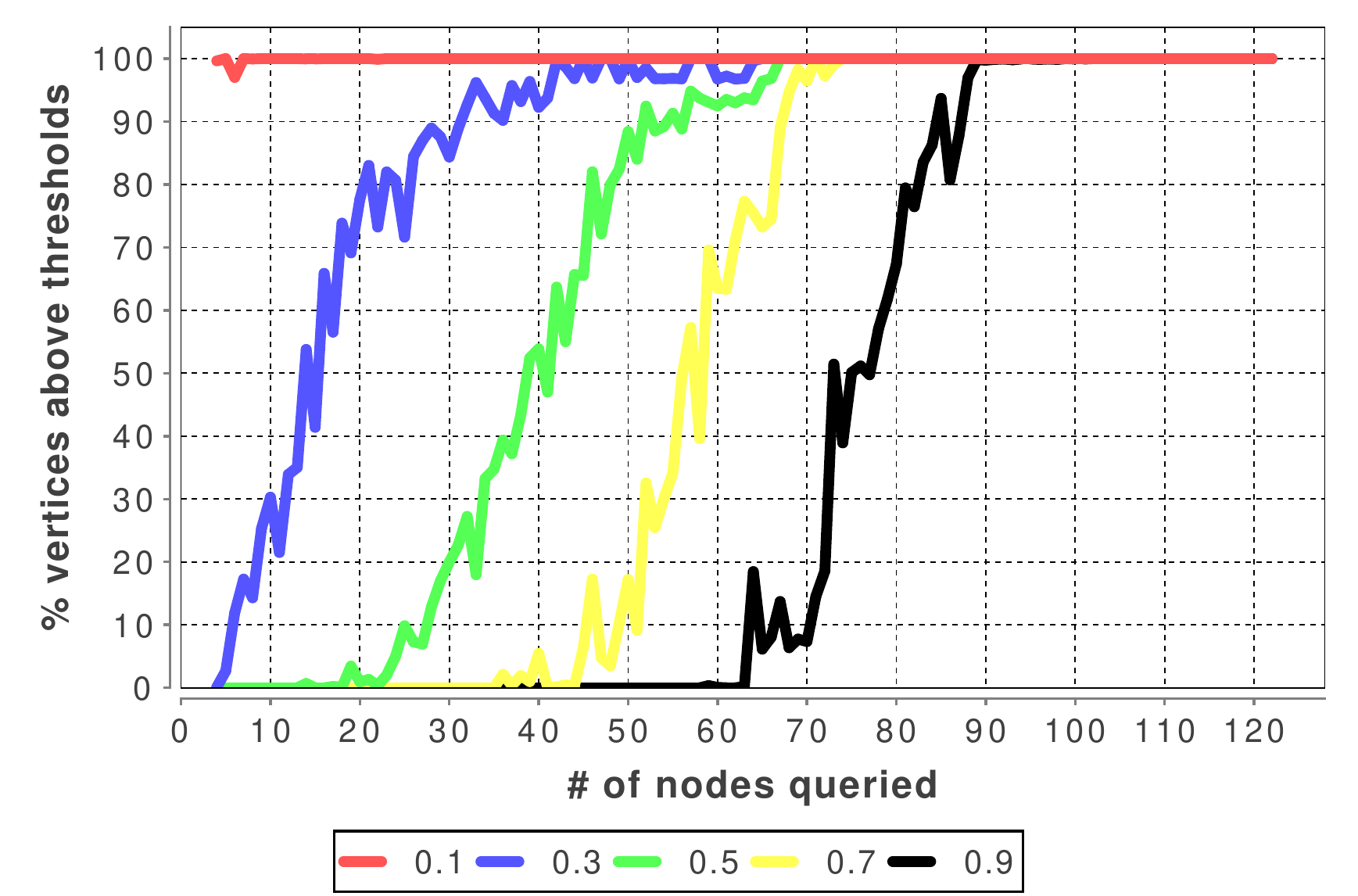}
\includegraphics[width=0.49\textwidth]{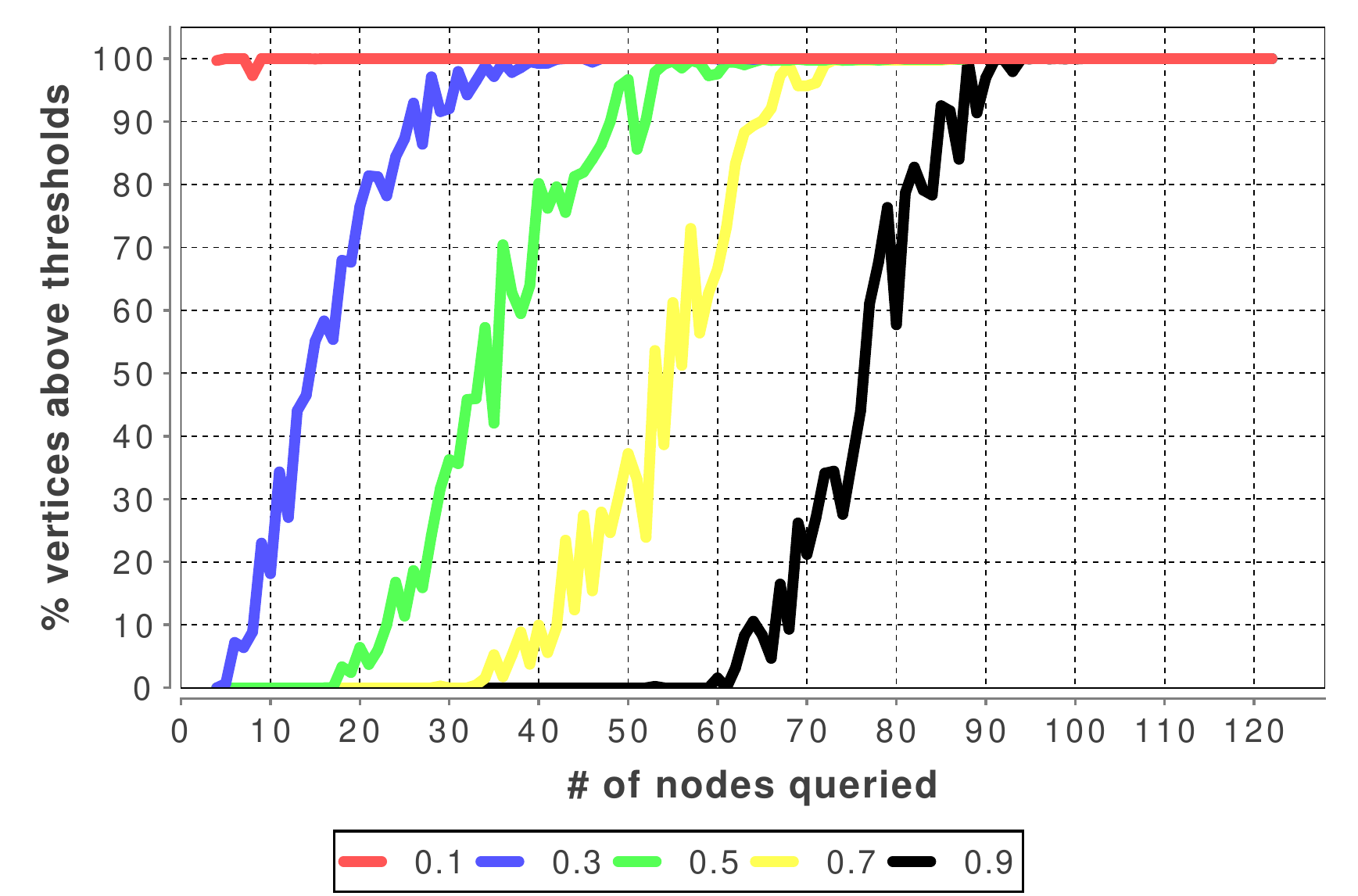}
\end{center}
\caption{Performance of the learning algorithms after the habitat attribute reassignment. With this new data set which better matches our block model, both learning algorithms achieve excellent performance. Left, MI; right, AA. }
\label{fig:habitat-fixpoint}
\end{figure}

In order to confirm our hypothesis that it is the accuracy of the block model, as opposed to the performance of the learning algorithm, that causes some vertices to be misclassified, we modified the data set in an artificial way in order to make it consistent with the block model.  Starting with the original data set, we iterated the following procedure: at each step, we assigned each species a new value of the habitat attribute, setting it equal to the most likely type according to the most likely block model, conditioned on the types of all other vertices.  After $6$ iterations of this process, changing the types of a total of $260$ species, we reached a fixed point, where the type of each vertex is consistent with the block model's predictions. As we expected, and shown in Fig.~\ref{fig:habitat-fixpoint}, our learning algorithms perform perfectly on this modified data set, predicting the type of every species with accuracy over 90\% after querying just 18\% of them.  

A direct interpretation of the query order on the Weddell Sea food web is difficult due to the complexity of the network. However, the query orders for the two different attributes have a lot in common, suggesting that they agree to a large extent about the relative importance of the species. As shown in Fig. 6, the query orders are positively correlated, with a Pearson’s coefficient of 0.553. The two attributes have a low correlation to begin with, as feeding types and habitats are close to orthogonal in ecosystems (species tend to fill the niches in the food chain wherever available). They have an Adjusted Mutual Information \cite{vinh_information_2009} of 0.357, which varies from 0 for a total lack of correlation (conditioned on the number of species of each type) and 1 for an exact match. As a result, we believe that the correlation between the query orders is largely due to the common underlying topology and its effect on the learning process.

\begin{figure}
\begin{center}
\includegraphics[width=0.69\textwidth]{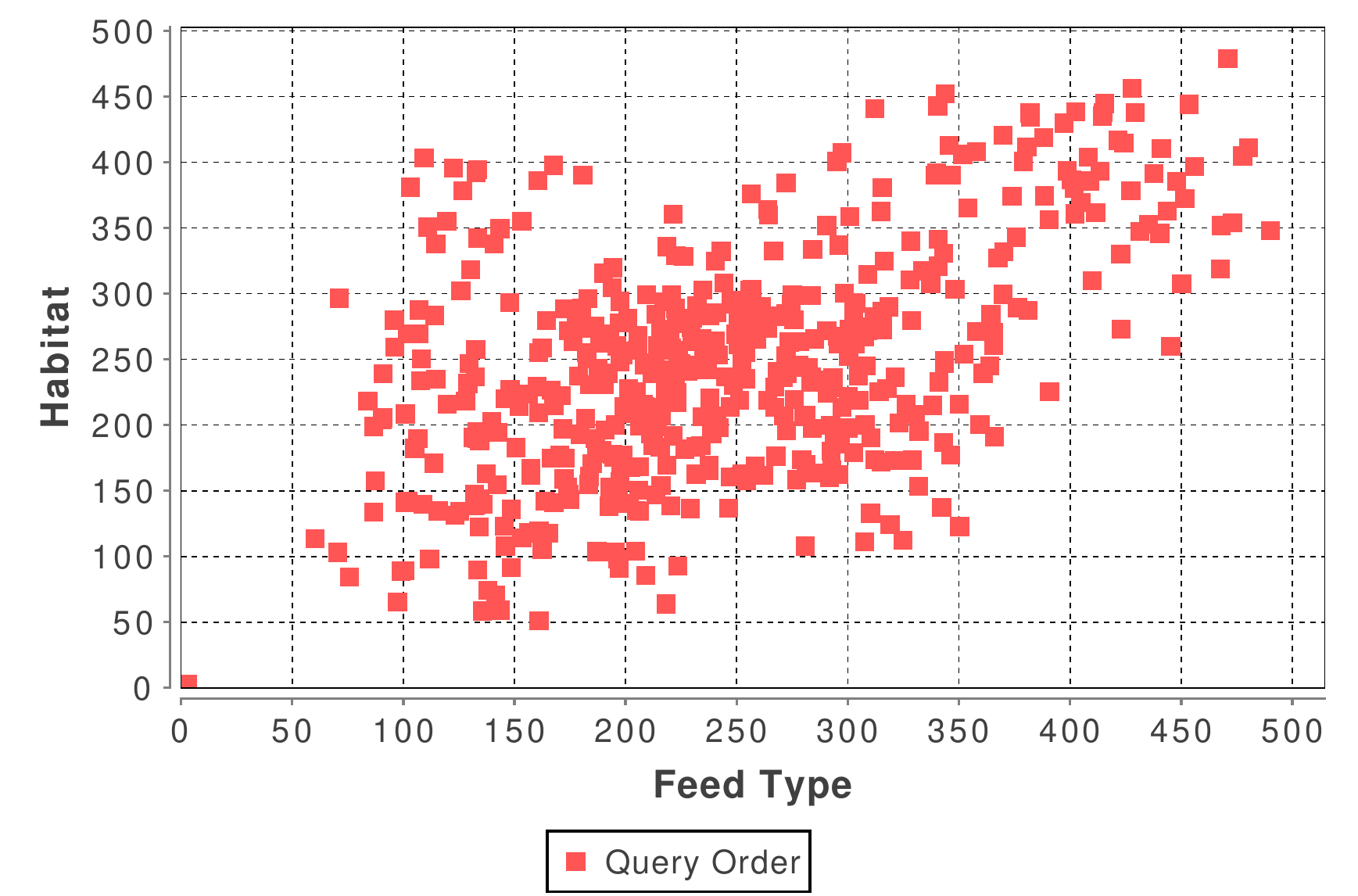}
\end{center}
\caption{The comparison between query orders for two different attributes on the Weddell Sea food web. $y$ axis: mean query order for the habitat attribute over 10 runs; $y$ axis: mean query order for the feeding type attribute over 10 runs. Data is from the same experiment shown in Fig.~\ref{fig:learnfoodweb}. The Pearson's coefficient between the query orders is 0.553, while the Adjusted Mutual Information between the attributes is 0.357.}
\label{fig:queryCompare}
\end{figure}   

\section{Comparison with Simple Heuristics}

To put these results into perspective, we compared our active learning algorithms with some simple heuristics. These include: 1) querying the vertex with highest degree in the subgraph of unqueried vertices, 2) querying the vertex with highest shortest path betweenness centrality \cite{brandes_faster_2001,newman_scientific_2001} in the subgraph of unqueried vertices and 3) querying a vertex uniformly at random from the unqueried ones. The first two heuristics are popular measures of centrality, which are believed to reflect the varying importance of the vertices in a network \cite{newman_measure_2005}. 

\begin{figure}
\begin{center}
\includegraphics[width=0.49\textwidth]{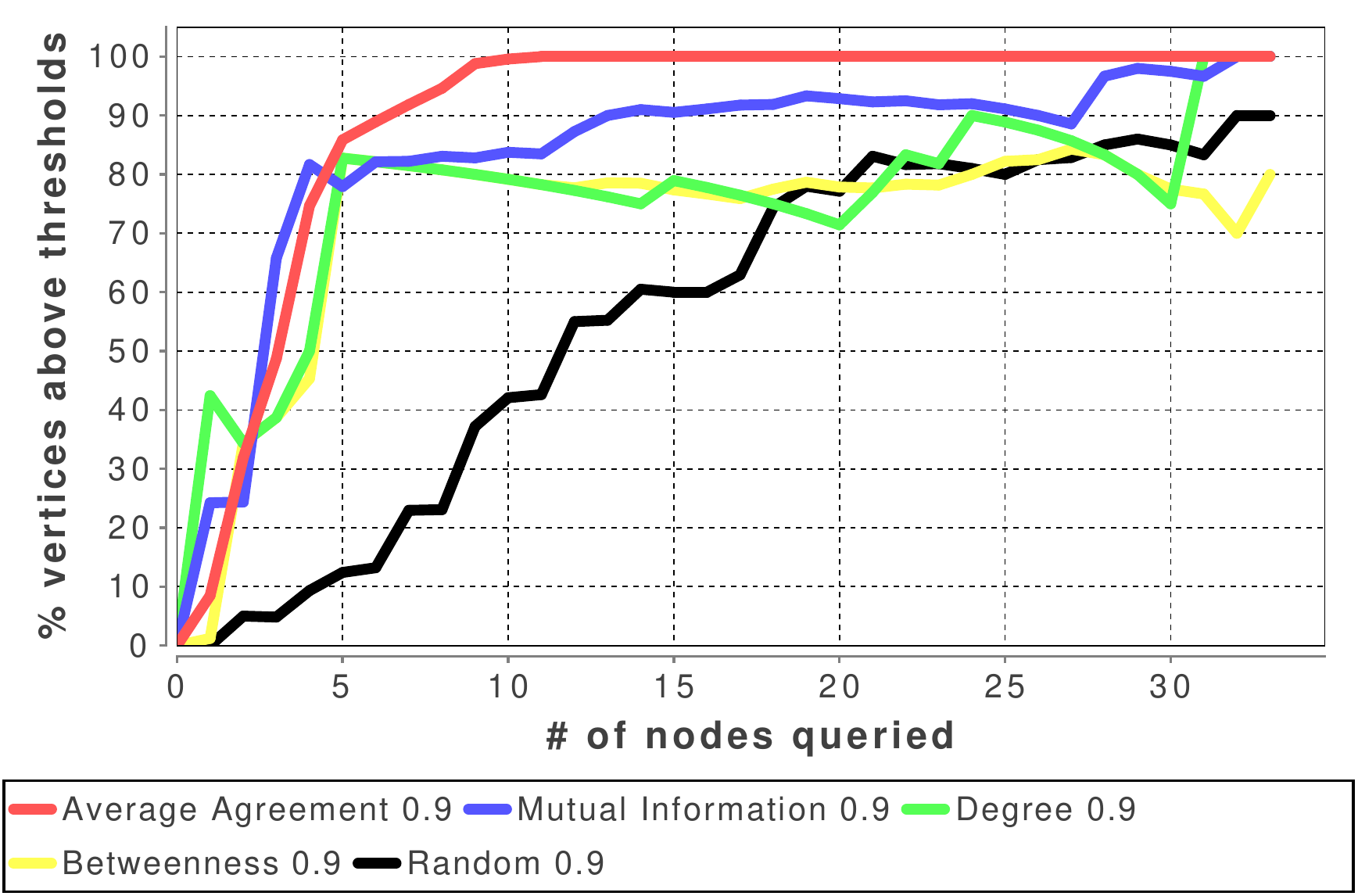}
\includegraphics[width=0.49\textwidth]{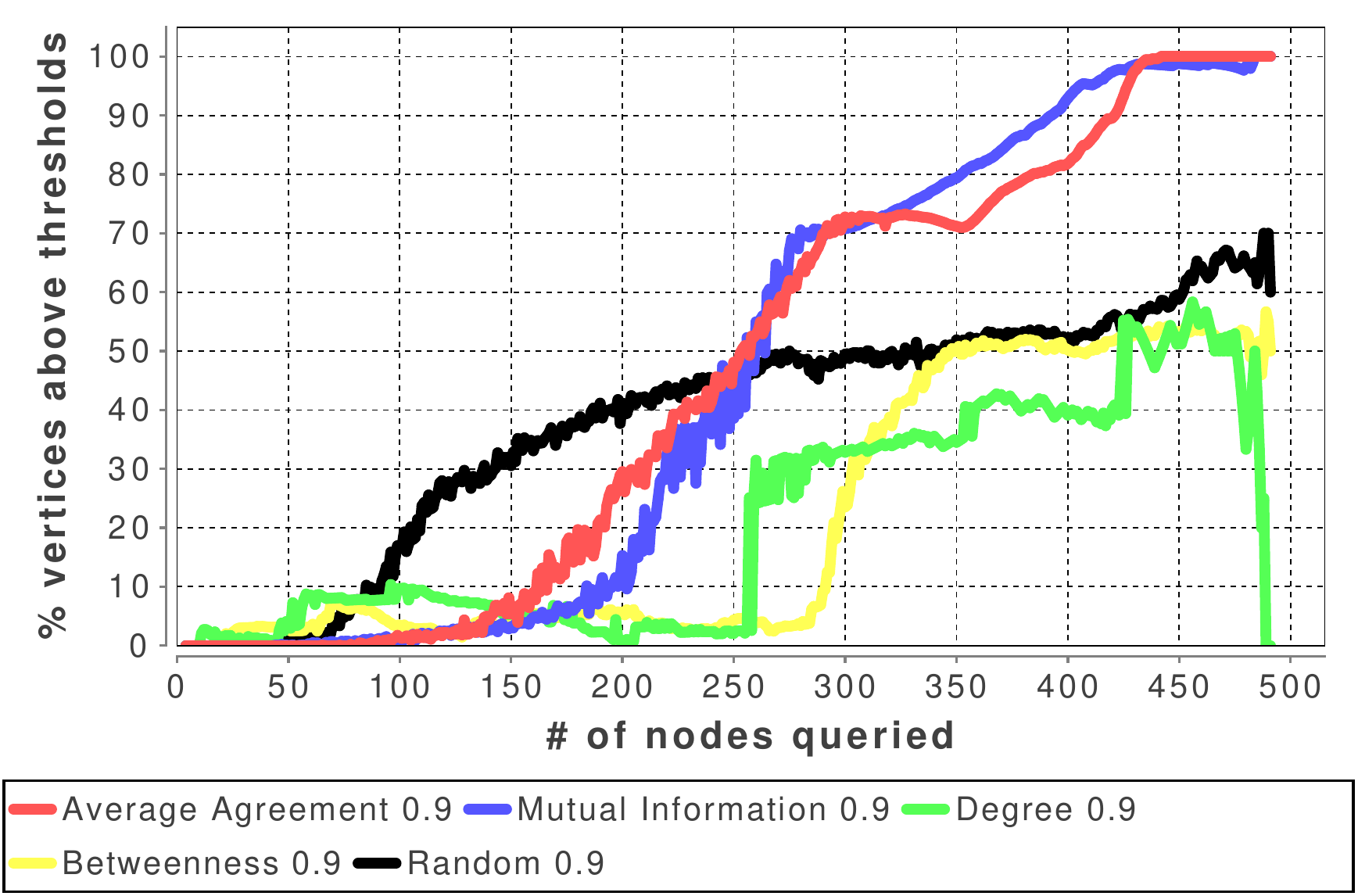}
\end{center}
\caption{A comparison of the MI and AA learning algorithms with three simple heuristics. Left, Zachary's Karate Club at $0.9$ accuracy threshold. Right, Weddell Sea food web with the feeding type attribute at $0.9$ accuracy threshold. All data are collected using the same Gibbs sampling process as specified in Fig.~\ref{fig:learnkarate} and Fig.~\ref{fig:learnfoodweb}.}
\label{fig:compare}
\end{figure}

We judge the performance of these heuristics using the same Gibbs sampling process for MI and AA. As Fig.~\ref{fig:compare} shows, on Zachary's Karate Club, although Degree and Betweenness did reasonably well, none of them beat MI or AA. For the Weddell Sea food web, however, the situation is more interesting. Random's curve still resembles a straight line, but its early performance is surprisingly good in comparison. We speculate that MI and AA need a burn-in process in the early stages of the process to achieve their full potential. Degree and Betweenness, on the other hand, did poorly throughout the process. It turns out some high degree/betweenness vertices are actually among the easiest to predict when rest the of graph is konwn. With some unpredictable low degree/betweenness vertices left unqueried, their accuracy remained quite low even when they had queried almost all the vertices.

\section{Conclusion}

Active learning, using mutual information and our average agreement measure, offers a new approach to dealing with networks where knowledge of vertex attributes is incomplete and costly.  Given that Gibbs sampling is computationally expensive, however, we do not expect the methods we used here to scale to truly large networks.  An interesting question is whether MI and AA can be estimated using other means, such as a message-passing algorithm---or where there are simple, scalable heuristics for selecting which vertex to query, based on some notion of betweenness or centrality, with similar performance.

In addition, the type of block model we use here does not deal well with sparse networks, or with heterogeneous degree distributions.  In particular, it tends to label high-degree and low-degree vertices as belonging to different types, with higher and lower values of $p_{ij}$.  In future work, we will test the MI and AA algorithms on a degree-corrected block model, such as in~\cite{morup_learning_????,_psift_genbio2008.pdf_????}, where the degrees of the vertices are part of the input, as opposed to data that the model is obliged to explain.


\subsubsection*{Acknowledgments}

We are grateful to Joel Bader, Aaron Clauset, Jennifer Dunne, Nathan Eagle, Brian Karrer, and Mark Newman for helpful conversations, and to Ute Jacob for the Weddell Sea food web data.  C.M. and J.-B. R. are also grateful to the Santa Fe Institute for their hospitality.  This work was supported by the McDonnell Foundation.

\bibliographystyle{splncs03}
\bibliography{reference}

\end{document}